\documentclass{article}

\usepackage{PRIMEarxiv}

\usepackage[utf8]{inputenc} 
\usepackage[T1]{fontenc}    
\usepackage{hyperref}       
\usepackage{url}            
\usepackage{booktabs}       
\usepackage{amsfonts}       
\usepackage{nicefrac}       
\usepackage{microtype}      
\usepackage{lipsum}
\usepackage{fancyhdr}       
\usepackage[table,xcdraw]{xcolor}
\usepackage{graphicx}
\usepackage{authblk}
\usepackage{forest}
\usepackage{caption}
\usepackage{subcaption}
\usepackage{amsmath}
\usepackage{diagbox}
\usepackage{dirtree}

\usepackage{todonotes}

\graphicspath{{media/}}     

\title{CESNET-TimeSeries24: Time Series Dataset for Network Traffic Anomaly Detection and Forecasting
}

\author[1, 2, *]{Josef Koumar}
\author[1]{Karel Hynek}
\author[1]{Tomáš Čejka}
\author[1]{Pavel Šiška}

\affil[1]{\footnotesize CESNET a.l.e., Generála Píky 430/26, 160 00 Prague 6, Czech Republic}
\affil[2]{\footnotesize Czech Technical University in Prague, Thákurova 9, 160 00 Prague 6, Czech Republic}
\affil[*]{corresponding author: Josef Koumar (josef.koumar@fit.cvut.cz)}

\begin{document}
\maketitle

\begin{abstract}
  


Anomaly detection in network traffic is crucial for maintaining the security of computer networks and identifying malicious activities. One of the primary approaches to anomaly detection are methods based on forecasting. Nevertheless, extensive real-world network datasets for forecasting and anomaly detection techniques are missing, potentially causing performance overestimation of anomaly detection algorithms. This manuscript addresses this gap by introducing a dataset comprising time series data of network entities' behavior, collected from the CESNET3 network. The dataset was created from 40 weeks of network traffic of 275 thousand active IP addresses. The ISP origin of the presented data ensures a high level of variability among network entities, which forms a unique and authentic challenge for forecasting and anomaly detection models. It provides valuable insights into the practical deployment of forecast-based anomaly detection approaches.

\end{abstract}


\section{Background \& Summary}


Traffic monitoring plays a crucial role in network management and overall computer security \cite{d2019survey}. Network-based intrusion detection and prevention systems can protect infrastructure against users' sloppiness, policy violations, or intentional attacks from inside. However, maintaining network security has become increasingly challenging due to the widespread adoption of traffic encryption, which significantly reduces visibility.

As a result, gaining insights into encrypted network traffic has become essential, particularly for threat detection. Recent research has focused on detecting security threats through the classification of encrypted traffic using machine learning techniques~\cite{aceto2019mobile,koumar2024nettisa,akbari2022traffic,luxemburk2023fine,plny2022decrypto}. However, in the domain of network traffic monitoring, there is a substantial challenge in acquiring up-to-date threat datasets~\cite{guerra2022datasets}.  Machine Learning classification models can detect already known attacks, which are captured in the dataset or those closely resembling them (such as malware from the same family). Therefore, unsupervised anomaly detection plays a crucial role in network traffic monitoring~\cite{yaacob2010arima} as it can identify unknown (zero-day) attacks due to behavioral changes caused by infection~\cite{andrysiak2014network}. 

The unsupervised anomaly detection method assigns anomalous scores to network entity behavior based on patterns and characteristics learned from historical data. One of the most used types of unsupervised anomaly detection algorithms is based on traffic forecasting (also referred as network traffic prediction). An anomaly alert is raised when the difference between the forecasted value and observation exceeds a defined threshold. Nevertheless, traffic forecasting can also be applied in other networking use cases, such as traffic management in data driven networks, resource allocation, and service orchestration.

In recent years, there has been a rapid development in forecasting and anomaly detection methods, not limited to computer science. Wu et al.~\cite{wu2021current} attributed this development to the rise and successful use of neural networks. Nevertheless, the recent performance improvement of forecasting methods applied to network traffic monitoring is uncertain due to the absence of long-term datasets~\cite{ferreira2023forecasting}. In their survey, Ferreira et al.~\cite{ferreira2023forecasting} describe the lack of a reference dataset as the crucial obstacle related to performance evaluation. Additionally, real-world datasets used in the evaluation are not publicly available due to privacy concerns. Therefore, the majority of publicly available datasets have synthetic origins. 

Synthetic data does not necessarily represent real-world tasks. Wu et al.~\cite{wu2021current} show that novel anomaly detection and forecasting approaches evaluated on synthetic datasets lead to the illusion of nonexisting progress. The more preferable option is to use up-to-date real-world data, like the MAWIlab project WIDE~\cite{cho2000traffic}, that publishes anonymized packet captures daily. However, the WIDE project provides only brief 15-minute daily packet traces~\cite{mawilab}, which is an insufficient time window for effective traffic modeling.

To address these challenges,  we decided to create a dataset called CESNET-TimeSeries24 that was collected by long-term monitoring of selected statistical metrics for 40 weeks for each IP address on the ISP network CESNET3 (Czech Education and Science Network). The dataset encompasses network traffic from more than 275,000 active IP addresses, assigned to a wide variety of devices, including office computers, NATs, servers, WiFi routers, honeypots, and video-game consoles found in dormitories. Moreover, the dataset is also rich in network anomaly types since it contains all types of anomalies identified by Chandola et al. and Basdekidou et al. \cite{chandola2009anomaly,basdekidou2017momentum}, ensuring a comprehensive evaluation of anomaly detection methods. Last but not least, the CESNET-TimeSeries24 dataset provides traffic time series on institutional and IP subnet levels to cover all possible anomaly detection or forecasting scopes. Overall, the time series dataset was created from the 66 billion IP flows that contain 4 trillion packets that carry approximately 3.7 petabytes of data. The CESNET-TimeSeries24 dataset is a complex real-world dataset that will finally bring insights into the evaluation of forecasting models in real-world environments.

\section{Methods} \label{sec:methods}
In this section, we provide detailed information about all methods used for obtaining the dataset. Since the dataset was obtained from a production network, CESNET3, and used by real users, privacy was a fundamental concern in our work,  leading us to conduct our research with careful consideration. The indisputable advantages of real traffic generated by hundreds of thousands of people come with understandable privacy concerns. Thus, we used only automatic data processing with immediate data anonymization. With this, we declare that we did not analyze or manually process non-anonymized data or perform any procedures that could allow us to track users or reveal their identities. 

The publication of the dataset has been approved by the Committee for Ethics in Research of the Scientific Council of the Czech Technical University in Prague under reference number 0000-10/24/51902/EKČVUT. The approval also includes a waiver of explicit user consent for publishing the dataset. Moreover, all users of the CESNET3 network agreed with the terms and conditions that define a monitoring process for optimization and improvement of services (including related research) and allow sharing of the data with third parties after anonymization (\url{https://www.cesnet.cz/en/gdpr}).





 
\subsection{Data capture}

\begin{figure*}
\centering
\includegraphics[width=13cm]{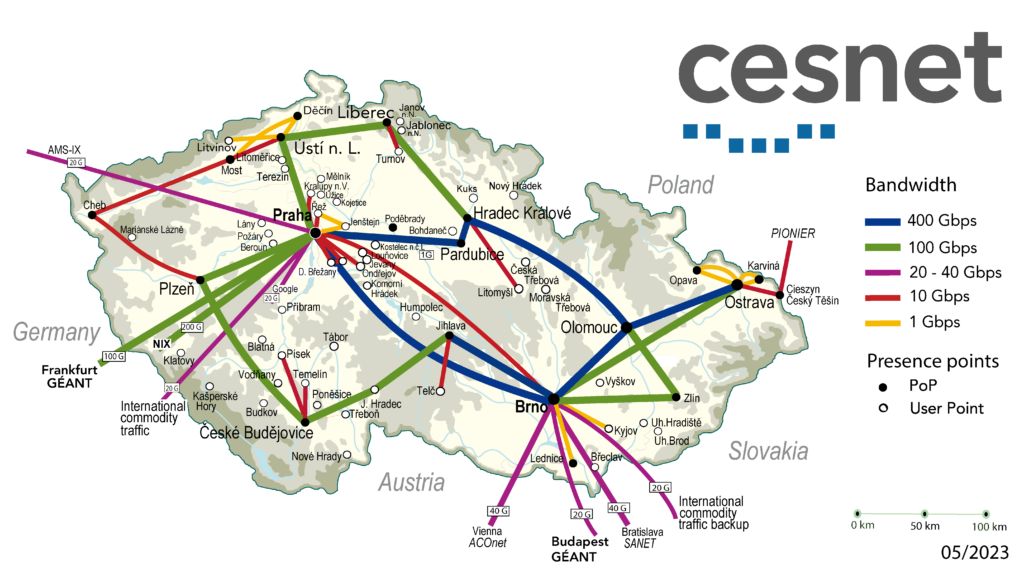}
\caption{Topology of the CESNET3 network, which interconnects academic institutions in the Czech Republic}
\label{fig:topology_cesnet2}
\end{figure*}

The network traffic was obtained from the CESNET3---an ISP network that provides internet access to public and research institutions in the Czech Republic. The network spans a whole country, as shown in Figure \ref{fig:topology_cesnet2}, and serves approximately half a million users daily. Since ISP networks transfer huge volumes of data, packet-based monitoring systems are infeasible due to the costs of processing and storage capacity. Therefore, the ISP networks (including CESNET3) are monitored using a standard IP flow monitoring system located at the perimeter---all transit lines to the peering partners are equipped with flow monitoring probes.


IP flow monitoring systems aggregate data packets into IP flow records. An IP flow record represents communication metadata associated with a single connection and is defined~\cite{rfc7011} as “a set of IP packets passing an observation point in the network during a certain time interval, such that all packets belonging to a particular flow have a set of common properties.” Commonly, these properties are referred to as flow keys and consist of source and destination IP addresses, transport layer ports, and protocol. 

The monitoring infrastructure for creating the dataset is detailed in Figure~\ref{fig:dataset_collection}. A network TAPs are positioned before the edge routers of the CESNET3 network and mirror the traffic to the monitoring probe, which is a server equipped with a high-speed monitoring card capable of handling 200\,GB/s. On the monitoring probe, the Ipfixprobe (\url{https://github.com/CESNET/ipfixprobe}) flow exporter is installed. The Ipfixprobe exporting process was set with an active timeout of 5 min and an inactive (idle) timeout of 65 s. Long connections are split when the connection duration is longer than the active timeout, and a flow record is exported even though the actual connection is not terminated yet. If no packet is observed within the inactive timeout period, the connection is considered terminated, and a flow record is exported. Using active and inactive timeouts for splitting connections is standard practice for flow-based network monitoring~\cite{rfc7011}. Additionally, the Ipfixprobe collected following features: start time, end time, number of packets, number of bytes, and Time to Live (TTL). None of the collected information contains application-level information, which ensures that the privacy of users' communication is not compromised. The collected data are then sent using IPFIX~\cite{rfc7011} protocol to the IP flow collector server with IPFIXcol2 (\url{https://github.com/CESNET/ipfixcol2}) flow collecting software installed.

\subsection{Time series aggregation}
\begin{figure*}[hb]
  \centering
  \includegraphics[width=\textwidth]{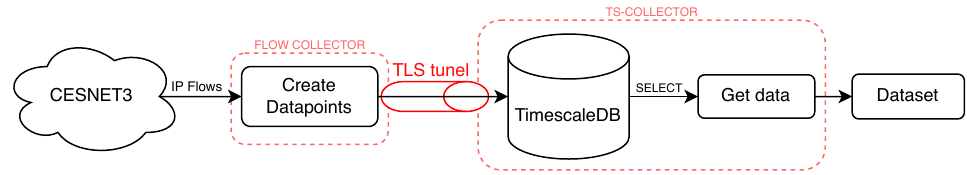}
  \caption{Architecture of dataset collection from the CESNET3 network}
  \label{fig:dataset_collection}
\end{figure*}

The flow collector server contains aggregation and filtration modules, as depicted in Figure~\ref{fig:dataset_collection}. The filtration module removes all transient traffic---where both source and destination addresses do not belong to CESNET3 and the packets are just passing through the network. Moreover, we also filter out all single TCP-SYN packet connections---scans. Given the ISP origin and probe placement before routers and firewalls, the scans would represent the absolute majority of the dataset. Moreover, a large number of scans would cause significant noise in the time series, which would result in possible false negatives in anomaly detection. Since the scans can be easily detected with simple methods~\cite{staniford2002practical,bhuyan2011surveying}, we decided to remove them from the dataset. All other flows are then passed to the aggregation module.


\begin{figure*}
  \centering
  \includegraphics[width=\textwidth]{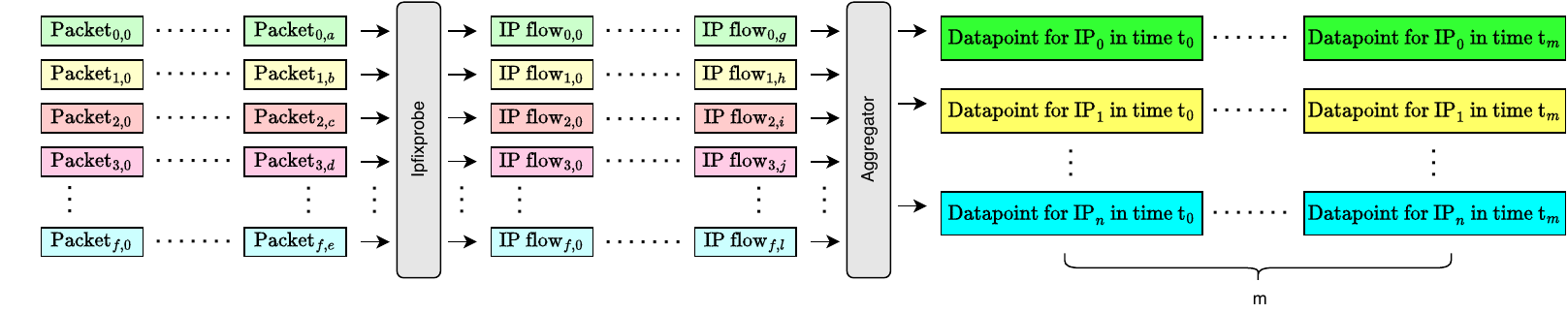}
  \caption{This diagram describes the aggregation process for capturing time series with length $ m $ from network traffic. For each packet $ Packet_{i,j} $ exists $ Flow_{i,k} $ where the packet belongs. Furthermore, it is always true that $ a \ge g $ (for $ b \ge h $ and others similarly), and in most cases, the $ a $ will be much larger than $ g $. Moreover, it is common that only one IP flow contains all packets from one connection ($ g = 0 $). This is common, for example, for connections generated by a user visiting a web page. Similarly, a time series datapoint is a combination of one or more IP flows.}
  \label{fig:traffic_preprocesing}
\end{figure*}

We generate evenly spaced time series for each IP address by aggregating IP flow records into time series datapoints. The process diagram for network traffic processing and aggregation is shown in Figure~\ref{fig:traffic_preprocesing}. The created datapoints represent the behavior of IP addresses within a defined time window of 10 minutes. The vector of time-series metrics $v_{ip, i}$ describes the IP address $ip$ in the $i$-th time window. Thus, IP flows for vector $v_{ip, i}$ are captured in time windows starting at $t_i$ and ending at $t_{i+1}$. The aggregated datapoints are then stored in TimeScaleDB (\url{https://www.timescale.com/}), where time series are built. 

Datapoints created by the aggregation of IP flows contain the following time-series metrics:
\begin{itemize}
  \item \textit{\textbf{Simple volumetric metrics:}} the number of IP flows, the number of packets, and the transmitted data size (i.e. number of bytes)
  \item \textit{\textbf{Unique volumetric metrics:}} the number of unique destination IP addresses, the number of unique destination Autonomous System Numbers (ASNs), and the number of unique destination transport layer ports. The aggregation of \textit{Unique volumetric metrics} is memory intensive since all unique values must be stored in an array. We used a server with 41 GB of RAM, which was enough for 10-minute aggregation on the ISP network.    
  \item \textit{\textbf{Ratios metrics:}} the ratio of UDP/TCP packets, the ratio of UDP/TCP transmitted data size, the direction ratio of packets, and the direction ratio of transmitted data size
  \item \textit{\textbf{Average metrics:}} the average flow duration, and the average Time To Live (TTL) 
\end{itemize}

\subsection{Anonymization}
The capturing process started on 9. October 2023 and ended on 14. July 2024. Thus, after 40 weeks, we extracted time series from the database. In the database framework, a script is deployed that automatically adds corresponding institution and institution subnet for each IP address. Nevertheless, we omit the extraction of exact IP addresses, institutions, and institution subnets. Instead, we used database IDs for IP addresses, institutions and institution subnets as an identifiers, which were randomly assigned during database creation. The omitting IP addresses, institutions and institution subnets in the data extraction performs effective anonymization. Nobody without access to the original database cannot revert this step and connect time series with a particular IP address, institution, or institutional network.

\subsection{Data preprocesing}

This subsection describes the preprocessing steps in detail. 

\textbf{Filtering: } The obtained raw dataset from the database contains a time series for approximately 400 thousand IP addresses. However, many of them were almost empty---network entities were not active most of the dataset creation time, resulting in an empty 10-minute aggregated record. Therefore, we remove time series with a smaller number of datapoints than 100, which is approximately $ 0.25 \%$ of the maximum datapoints that the time series in this dataset can contain. This action results in a time series for 275,124 IP addresses.


\textbf{Multiple time aggregation: } The original datapoints in the dataset are aggregated by 10 minutes of network traffic. The size of the aggregation interval influences anomaly detection procedures, mainly the training speed of the detection model. However, the 10-minute intervals can be too short for longitudinal anomaly detection methods. Therefore, we added two more aggregation intervals to the datasets. 

We provide additional one-hour and one-day aggregation intervals. These aggregated intervals were created from the 10-minute interval. When possible, time series metrics were aggregated using the sum of values (such as IP flows, number of packets and transimitted bytes). Nevertheless, metrics that represent a number of unique values cannot be easily summed without losing potentially important information. Therefore, in that case, we provide sum, mean, and standard deviation that results in three new time series metrics per each original metric. Finally, we use the mean for the time series metrics, which represent ratio or average values. 

\textbf{Time series of institutions: } Many security events can be visible only from the perspective of overall network traffic. Therefore, we use the institution ID exported from the database to divide the IP addresses into groups based on institutions. We identify 283 institutions inside the CESNET3 network. These time series aggregated per each institution ID provide a view of the institution's data. 

\textbf{Time series of institutional subnets: } Many institutions have multiple networks, which are usually in different locations or have different purposes in the organization. The administrators usually like to handle security not only overall but also per each subnet. Therefore, we divide the IP addresses into groups based on institution subnets in the same location using information from ISP CESNET. These time series aggregated per each institution's network provide a view of data that would probably be monitored by the institution's SoC team. 

We identify 548 institution subnets inside the CESNET3 network. Almost 75 \% of institutions have exactly one network. Moreover, the next 14 \% of institutions have two networks. However, there is also an institution with 105 networks. This is because the CESNET3 network interconnects to a lot of high schools, hospitals, and museums. Therefore, 75 \% of institutions' time series is in the institution subnets' time series dataset.

\textbf{Weekend and holidays: } In the network, traffic forecasting and anomaly detection can be crucial to add information about weekends and holidays into the model training and evaluation. Therefore, we included weekend dates and dates of public holidays in the Czech Republic for convenience. 


\DTsetlength{0.3em}{1em}{0.2em}{0.5pt}{2.5pt}
\setlength{\DTbaselineskip}{7pt}
\begin{figure}[ht!]
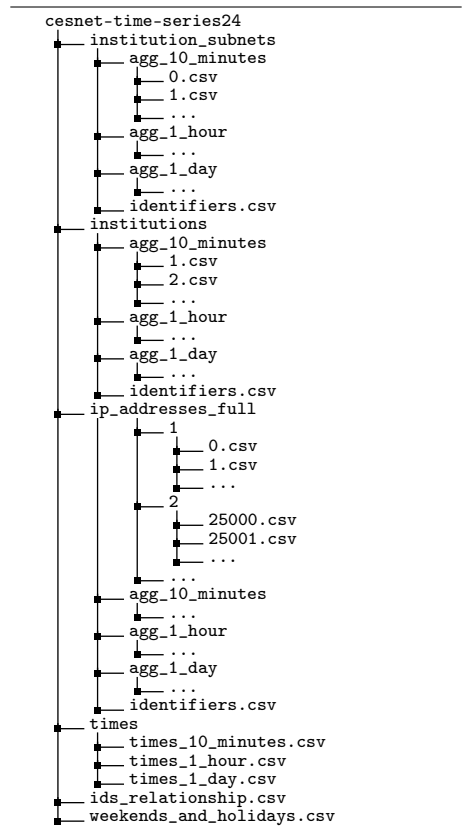

\centering
\begin{minipage}{6cm}
\noindent\rule{\textwidth}{0.4pt}
\scriptsize
\dirtree{%
    .1 cesnet-time-series24.
    .2 institution\_subnets.
    .3 agg\_10\_minutes.
    .4 0.csv.
    .4 1.csv.
    .4 \ldots.
    .3 agg\_1\_hour.
    .4 \ldots.
    .3 agg\_1\_day.
    .4 \ldots.
    .3 identifiers.csv.
    .2 institutions.
    .3 agg\_10\_minutes.
    .4 1.csv.
    .4 2.csv.
    .4 \ldots.
    .3 agg\_1\_hour.
    .4 \ldots.
    .3 agg\_1\_day.
    .4 \ldots.
    .3 identifiers.csv.
    .2 ip\_addresses\_full.
    .4 1.
    .5 0.csv.
    .5 1.csv.
    .5 \ldots.
    .4 2.
    .5 25000.csv.
    .5 25001.csv.
    .5 \ldots.
    .4 \ldots.
    .3 agg\_10\_minutes.
    .4 \ldots.
    .3 agg\_1\_hour.
    .4 \ldots.
    .3 agg\_1\_day.
    .4 \ldots.
    .3 identifiers.csv.
    .2 times.
    .3 times\_10\_minutes.csv.
    .3 times\_1\_hour.csv.
    .3 times\_1\_day.csv.
    .2 ids\_relationship.csv.
    .2 weekends\_and\_holidays.csv.
}

\noindent\rule{\textwidth}{0.4pt}
\end{minipage}
\caption{The file structure of the CESNET-TimeSeries24 dataset.}
\label{fig:dataset-dirtree}
\end{figure}

\section{Data Records} \label{sec:data_records}
The dataset is delivered in the form of compressed CSV files and is available at the Zenodo Platform via URL link: \url{https://zenodo.org/records/13382427}. This section provides the dataset structure and explains all data fields and files. 

\subsection{Data file hierarchy}
The dataset structure is outlined in Figure~\ref{fig:dataset-dirtree}. Each time series type (either institutional subnets, institutions, or raw IP addresses) is also divided by the aggregation window (10 minutes, 1 hour, and 1 day). The time series data, containing all relevant features, is stored in CSV files. Each file is named according to the identifier of the entity whose behavior the time series follows, ensuring a clear association between the time series and the corresponding entity. Since identifiers are not a row of consecutive numbers due to previous filtration and some numbers are missing, each time series type has its own \texttt{identifiers.csv} file, which contains lists of all identifiers of the included entities.

The time series representing the activity of individual IP addresses is additionally divided into multiple subdirectories. Since there are 275,124 IP address time series in the dataset, each located in its own file, some filesystems had difficulty handling such a large number of files in a single directory. Therefore, we organized them into subdirectories, each with 25,000 files. For individual IP addresses, the \texttt{identifiers.csv} file also includes the name of the subdirectory where the corresponding time series data is stored.

The dataset also contains a times directory, which contains translation tables between the sequence number of the aggregation time window and the absolute time. Such translation is often useful for plots with forecasting performance. Finally, it also contains a relationship table between IP addresses, institutions, and institution subnets, as well as a table denoting public holidays and weekends.

 \subsection{File format}
 The dataset files are organized as tables in CSV file format. There are two different file formats for the time series. The time series aggregated over 10 minutes contain columns described in Table \ref{tab:aggregation_metrics}. The time series aggregated over one hour and one day contains additional data features due to reaggregation. Therefore, apart of features described in Table~\ref{tab:aggregation_metrics} it also contains features described in Table~\ref{tab:aggregation_metrics_2}. Moreover, the dataset contains two support files--\texttt{ids\_relationship.csv} and \texttt{weekends\_and\_holidays.csv}. Table~\ref{tab:id_relationship} describes the content of \texttt{ids\_relationship.csv}, which provides information about the relationship between the IP addresses, institutions, and institution subnets dataset types. Furthermore, Table~\ref{tab:weekends_and_holidays} describes the content of \texttt{weekends\_and\_holidays.csv} that provides information about which days are non-working days in the Czech Republic (weekends and national holidays). 
        
      \begin{table*}
        \centering
        \caption{\label{tab:aggregation_metrics}Detailed descriptions of time series metrics for each IP address dataset.}
        \begin{tabular}{m{4cm}|m{12cm}}
            \textbf{Column name } & \textbf{Description} \\
            \hline
            \texttt{id\_time} & Unique identifier for each aggregation interval within the time series, used to segment the dataset into specific time periods for analysis. \\
            \rowcolor[HTML]{EFEFEF} \texttt{n\_flows} & Total number of flows observed in the aggregation interval, indicating the volume of distinct sessions or connections for the IP address. \\
            \texttt{n\_packets} & Total number of packets transmitted during the aggregation interval, reflecting the packet-level traffic volume for the IP address. \\
            \rowcolor[HTML]{EFEFEF} \texttt{n\_bytes} & Total number of bytes transmitted during the aggregation interval, representing the data volume for the IP address. \\
            \texttt{n\_dest\_ip} & Number of unique destination IP addresses contacted by the IP address during the aggregation interval, showing the diversity of endpoints reached. \\
            \rowcolor[HTML]{EFEFEF} \texttt{n\_dest\_asn} & Number of unique destination Autonomous System Numbers (ASNs) contacted by the IP address during the aggregation interval, indicating the diversity of networks reached. \\
            \texttt{n\_dest\_port} & Number of unique destination transport layer ports contacted by the IP address during the aggregation interval, representing the variety of services accessed. \\
            \rowcolor[HTML]{EFEFEF} \texttt{tcp\_udp\_ratio\_packets} & Ratio of packets sent using TCP versus UDP by the IP address during the aggregation interval, providing insight into the transport protocol usage pattern. This metric belongs to the interval <0, 1> where 1 is when all packets are sent over TCP, and 0 is when all packets are sent over UDP. \\
            \texttt{tcp\_udp\_ratio\_bytes} & Ratio of bytes sent using TCP versus UDP by the IP address during the aggregation interval, highlighting the data volume distribution between protocols. This metric belongs to the interval <0, 1>  with same rule as \texttt{tcp\_udp\_ratio\_packets} \\
            \rowcolor[HTML]{EFEFEF} \texttt{dir\_ratio\_packets} & Ratio of packet directions (inbound versus outbound) for the IP address during the aggregation interval, indicating the balance of traffic flow directions. This metric belongs to the interval <0, 1>, where 1 is when all packets are sent in the outgoing direction from the monitored IP address, and 0 is when all packets are sent in the incoming direction to the monitored IP address. \\
            \texttt{dir\_ratio\_bytes} & Ratio of byte directions (inbound versus outbound) for the IP address during the aggregation interval, showing the data volume distribution in traffic flows. This metric belongs to the interval <0, 1> with the same rule as \texttt{dir\_ratio\_packets}. \\
            \rowcolor[HTML]{EFEFEF} \texttt{avg\_duration} & Average duration of IP flows for the IP address during the aggregation interval, measuring the typical session length. \\
            \texttt{avg\_ttl} & Average Time To Live (TTL) of IP flows for the IP address during the aggregation interval, providing insight into the lifespan of packets. \\
            \hline
        \end{tabular}
      \end{table*}

        \begin{table*}
        \centering
        \caption{\label{tab:aggregation_metrics_2}Time series metrics which replace metrics \texttt{n\_dest\_ip}, \texttt{n\_dest\_asn} and \texttt{n\_dest\_port} in aggregation.}
        \begin{tabular}{p{4cm}|p{12cm}}
            \textbf{Column name } & \textbf{Description} \\
            \hline
            \texttt{sum\_n\_dest\_ip} & Sum of numbers of unique destination IP addresses. \\
            \rowcolor[HTML]{EFEFEF} \texttt{avg\_n\_dest\_ip} & The average number of unique destination IP addresses. \\
            \texttt{std\_n\_dest\_ip} & Standard deviation of numbers of unique destination IP addresses. \\
            \rowcolor[HTML]{EFEFEF} \texttt{sum\_n\_dest\_asn} & Sum of numbers of unique destination ASNs. \\
            \texttt{avg\_n\_dest\_asn} & The average number of unique destination ASNs. \\
            \rowcolor[HTML]{EFEFEF} \texttt{std\_n\_dest\_asn} & Standard deviation of numbers of unique destination ASNs) \\
            \texttt{sum\_n\_dest\_port} & Sum of numbers of unique destination transport layer ports. \\
            \rowcolor[HTML]{EFEFEF} \texttt{avg\_n\_dest\_port} & The average number of unique destination transport layer ports. \\
            \texttt{std\_n\_dest\_port} & Standard deviation of numbers of unique destination transport layer ports. \\
            \hline
        \end{tabular}

        \vspace{0.1cm}

        \centering
        \caption{\label{tab:id_relationship}Content of the \texttt{ids\_relationship.csv} file}
        \begin{tabular}{p{4cm}|p{12cm}}
            \textbf{Column name } & \textbf{Description} \\
            \hline
            \texttt{id\_ip} & ID of the IP address  \\
            \rowcolor[HTML]{EFEFEF} \texttt{id\_institution} & ID of the institution which own the IP address \\
            \texttt{id\_institution\_subnet} & ID of the institution subnet in which the IP address belongs \\
            \hline
        \end{tabular}

        \vspace{0.1cm}

        \centering
        \caption{\label{tab:weekends_and_holidays}Content of the \texttt{weekends\_and\_holidays.csv} file}
        \begin{tabular}{p{4cm}|p{12cm}}
            \textbf{Column name } & \textbf{Description} \\
            \hline
            \texttt{Date} & The date of the day in format "YYYY-MM-HH" \\
            \rowcolor[HTML]{EFEFEF} \texttt{Type} & The type of the day (Weekend or Holiday) \\
            \hline
        \end{tabular}
      \end{table*}

\section{Technical Validation} \label{sec:technical_validation}


This section provides technical validation of the dataset and is divided into three parts: 1) Validation of overall dataset properties, 2) Validation of the existence of anomalies, and 3) Validation of usability of the dataset for forecasting approaches.

\subsection{Validation of overall dataset properties}

In this section, we aim to validate the overall statistical properties of the dataset across the 40 weeks. 

\begin{figure*}
  \centering
  \includegraphics[width=\textwidth]{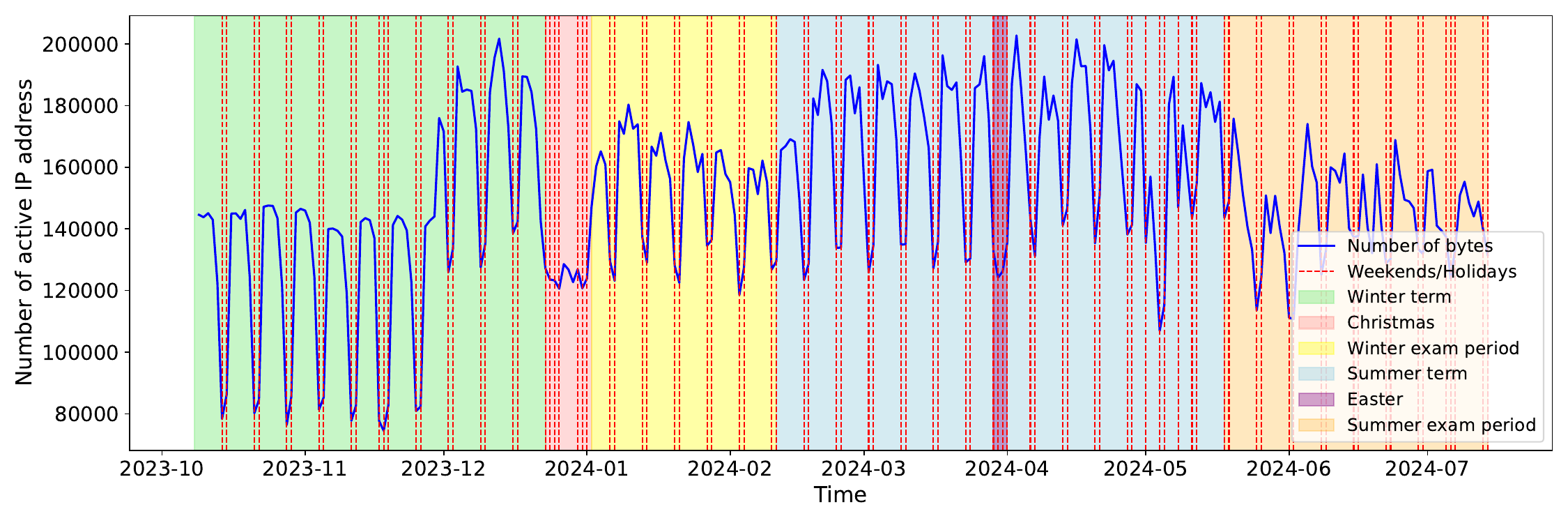}
  \caption{The evolution of the number of active IP addresses for each dataset's day }
  \label{fig:overall_ip_activity}
\end{figure*}

\paragraph{Activity of IP addresses} The dataset contains network traffic of more than 275 thousand IP addresses. However, these IP addresses typically do not communicate constantly over time. The evolution of the number of active IP addresses for each dataset's day is shown in Figure~\ref{fig:overall_ip_activity}. We can see that the number of active IP addresses correlates with the weekends and holidays, which is highly expected behavior. Moreover, we can see a slight correlation with terms and exam periods, which is caused by the CESNET3 interconnecting many universities and dormitories.

\begin{figure*}
    \centering
    \begin{subfigure}[b]{\textwidth}
        \centering
        \includegraphics[width=\textwidth]{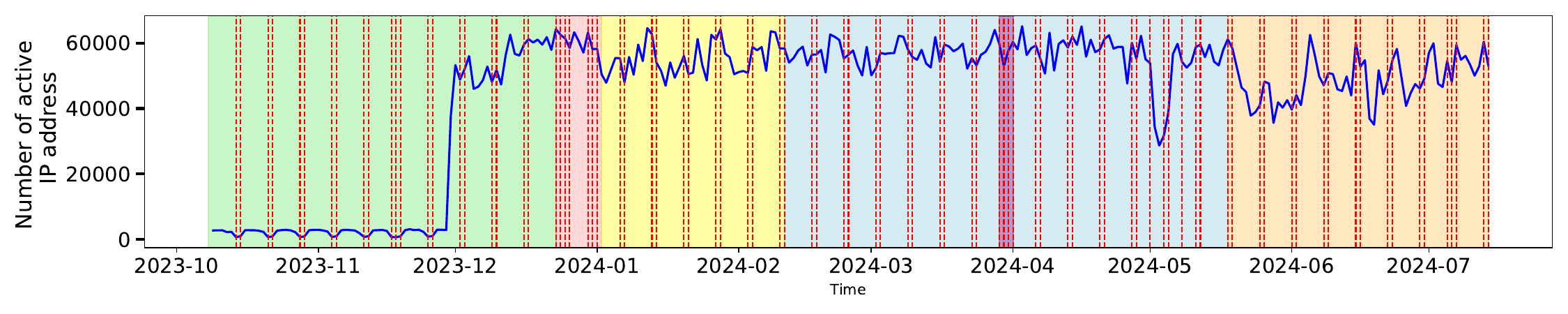}
        \caption{The evolution of the number of active problematic university's IP addresses for each dataset's day}
        \label{fig:problematic_institution-upper}
    \end{subfigure}

    \begin{subfigure}[b]{\textwidth}
        \centering
        \includegraphics[width=\textwidth]{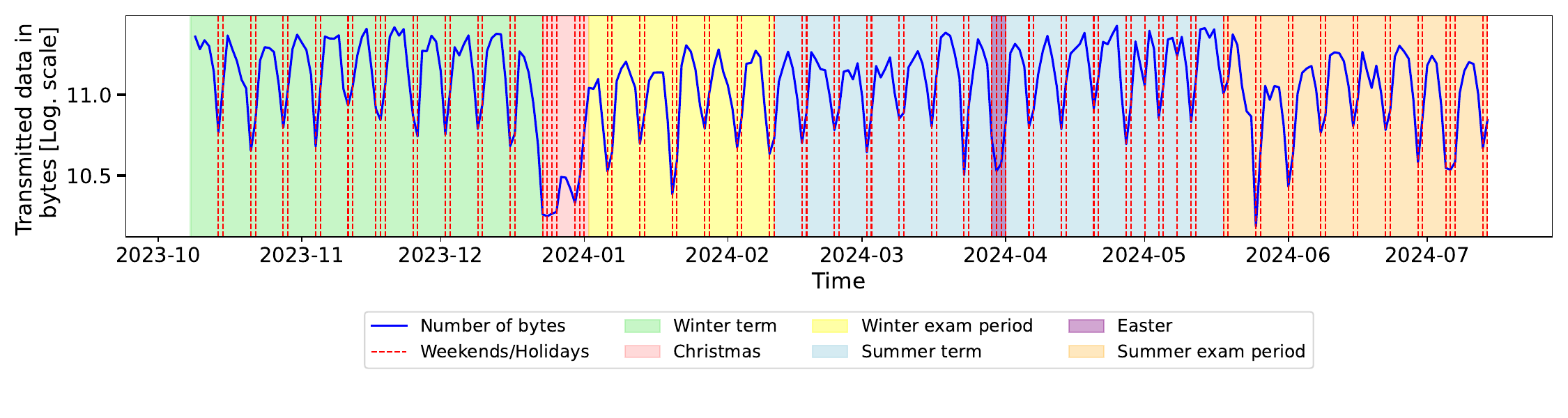}
        \caption{The evolution of overall transmitted data by problematic university}
        \label{fig:problematic_institution-lower}
    \end{subfigure}

    \caption{The comparison of evolution of active IP addresses and transmitted data by problematic university}
    \label{fig:problematic_institution}
\end{figure*}

However, the number of active IP addresses increased by approximately 50 thousand IP addresses at the end of November. We further evaluated this anomaly and found out it was caused by a single institution, which is one of the universities connected to the CESNET3 network. The evolution of the number of active IP addresses belonging to this university is shown in Figure \ref{fig:problematic_institution-upper}. We can see that the number of IP addresses significantly increased at the end of November. However, the evolution of transmitted data does not correlate with the evolution of a number of active IP addresses which can be seen in Figure \ref{fig:problematic_institution-lower}.

We discussed our findings with experts in CESNET. We provided them with the newly occurring IDs of IP addresses, and they found out that this anomaly is caused by the change in the university's network architecture, which, from the end of November, used public IP addresses for Eduroam WiFi.

\begin{figure*}
  \centering
  \includegraphics[width=\textwidth]{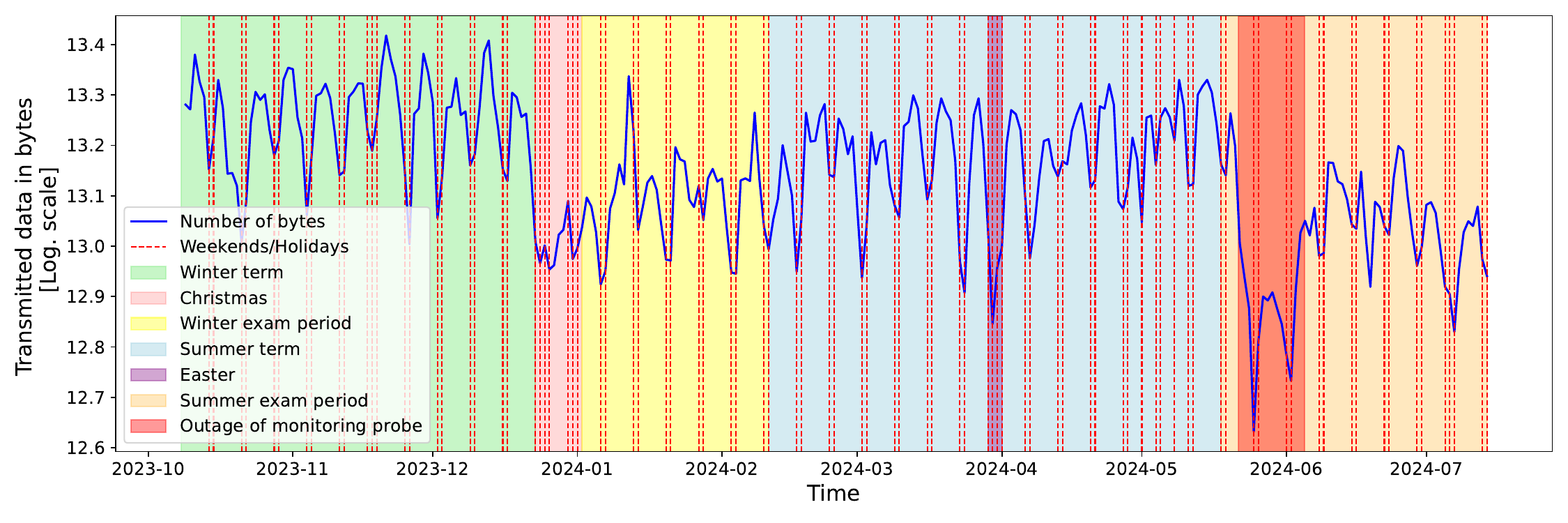}
  \caption{Overall data that were transmitted on the CESNET3 network and are captured in the dataset}
  \label{fig:overall_traffic_load}
\end{figure*}

\paragraph{Evolution of transmitted data} Overall data that were transmitted by IP addresses in the dataset is shown in Figure \ref{fig:overall_traffic_load}. The transmitted data in the aggregation window is stored in the time series metrics \texttt{n\_bytes}. In the figure, it can be seen that the evolution of transmitted data size highly correlates with weekends and holidays. Moreover, CESNET3 interconnects universities and dormitories; thus, we can also see a correlation between the terms and the exam periods. This observation is in line with the findings of Beneš et al. \cite{benes2023look}.

Furthermore, it can be seen that the decrease in traffic after the end of the summer term is even larger than the decrease in traffic during Christmas. Therefore, we performed the evaluation of this anomaly, and we found out that one of the monitoring probes was broken from approximately \texttt{2024.05.21 16:30} to approximately \texttt{2024.06.04 20:00}. So, the probe did not send IP flows to the IP flow collector.

\paragraph{Gaps in time series} Real-world network traffic data often contains gaps where a device does not transmit any data. In some instances, the entire network's traffic may exhibit such gaps. These gaps, or spaces, pose a challenge that must be addressed before applying forecasting methods.

One way to manage these gaps is through the aggregation process. If the aggregation window is sufficiently large, the resulting time series might eliminate these gaps. However, for scenarios involving multiple processes, such as traffic from multiple devices or entire networks, it's unlikely that a single aggregation window will be effective across the board. Additionally, using a large aggregation window can obscure important patterns in the time series, such as anomalies. As a result, it's inevitable that some time series will contain gaps.

\begin{figure*}
    \centering
    \begin{subfigure}[b]{0.33\textwidth}  
        \centering
        \includegraphics[width=\textwidth]{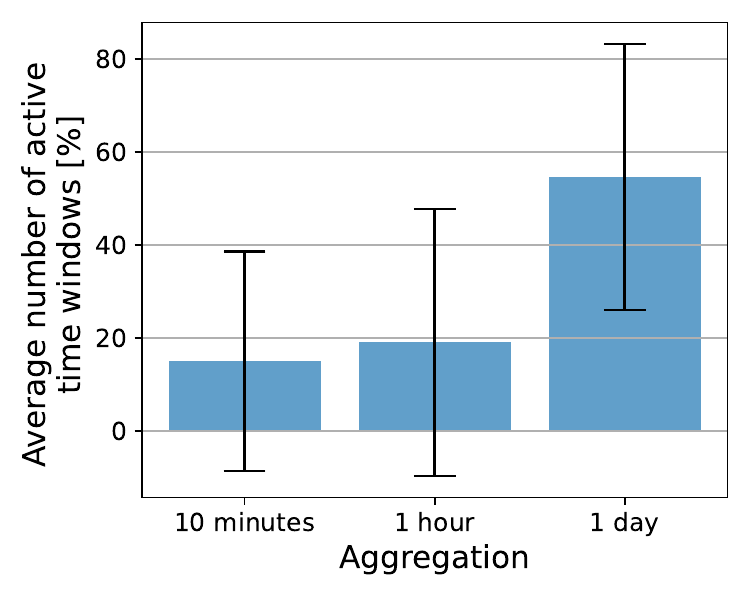}
        \caption{Analysis of the average number of active aggreagation windows}
        \label{fig:zeros-sub1}
    \end{subfigure}
    \hfill
    \begin{subfigure}[b]{0.66\textwidth}  
        \centering
        \includegraphics[width=\textwidth]{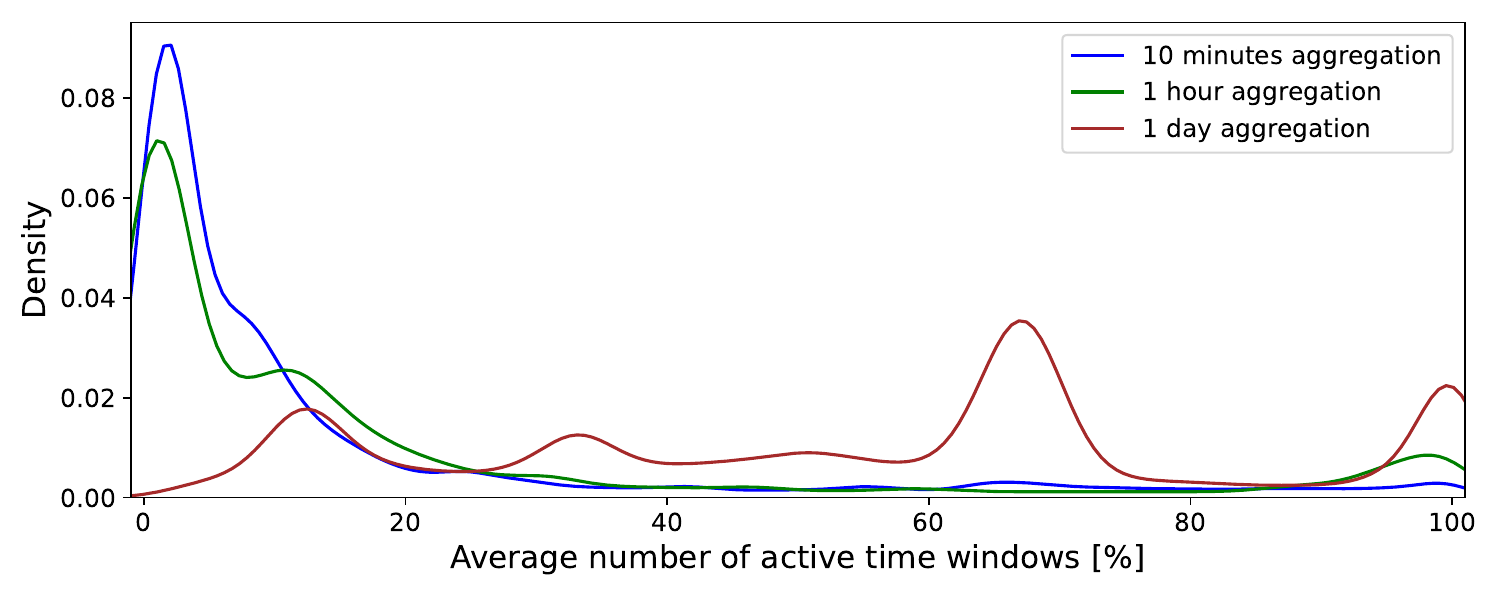}
        \caption{Analysis of active time windows using Kernel Density Estimation (KDE) density function}
        \label{fig:zeros-sub2}
    \end{subfigure}
    \caption{Analysis of gaps in dataset}
    \label{fig:zeros}
\end{figure*}

In our dataset, many time series contain significant gaps. In Figure \ref{fig:zeros-sub1} shows the average percentage of active datapoints in these time series, along with the standard deviation. For the 10-minute aggregation interval, active datapoints make up less than 20\% of the time series, meaning that gaps constitute more than 80\% on average. As expected, the number of gaps decreases as the aggregation interval increases. However, even with a day-long aggregation interval, gaps still account for more than 40\% of the time series on average.

Further analysis reveals an additional distribution pattern. The Kernel Density Estimation (KDE) distribution function in Figure \ref{fig:zeros-sub2} highlights this, particularly during the day aggregation, where there is a peak between 60\% and 70\%. This peak corresponds to the percentage of working days in the dataset, which is 67.5\%. This suggests that a significant portion of the time series represents workstations, as expected.


\subsection{Validation of existence of anomalies}

\begin{figure*}
  \centering
  \includegraphics[width=\textwidth]{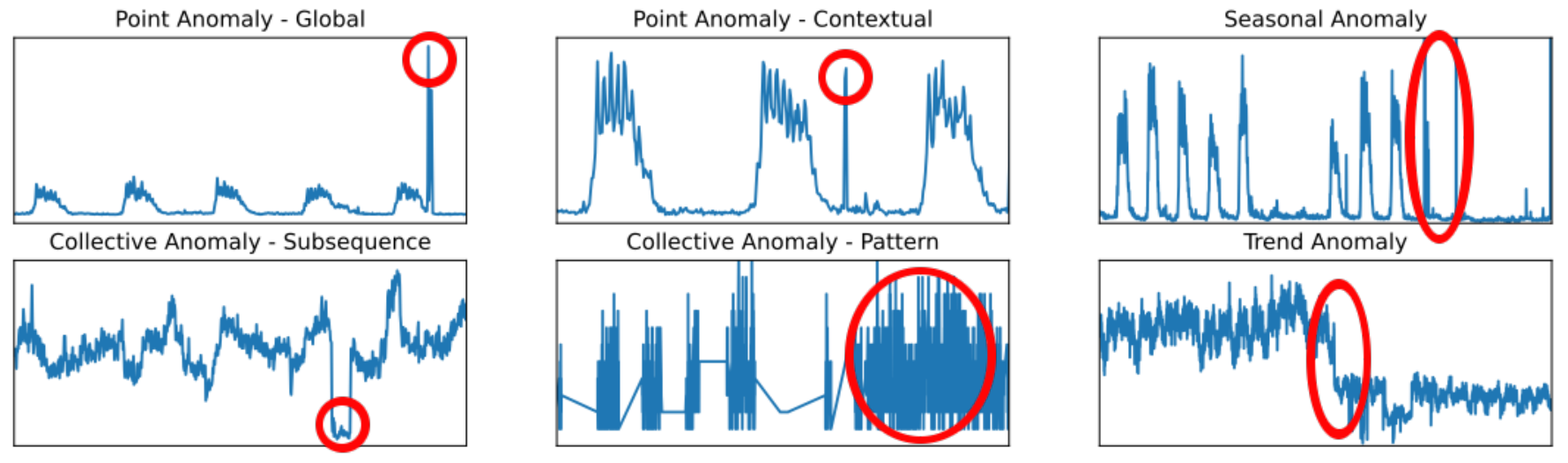}
  \caption{Examples of the anomaly types occurring in the dataset}
  \label{fig:anomaly_types}

  \centering
  \includegraphics[width=\textwidth]{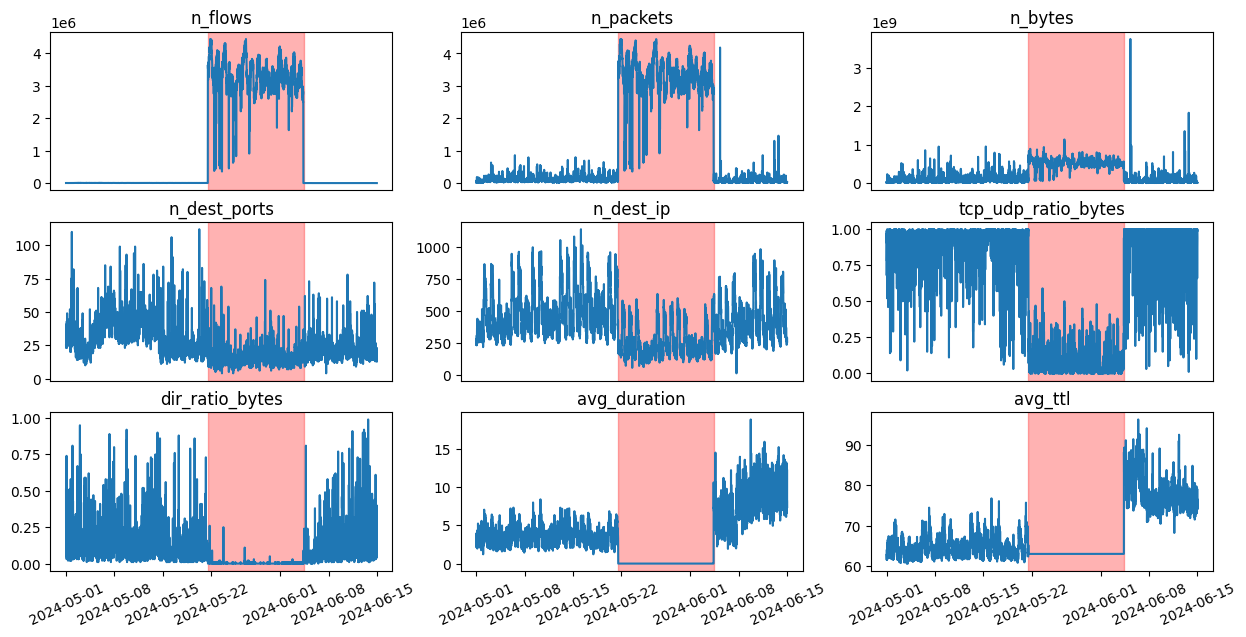}
  \caption{Analysis of time series metrics of detected anomaly on time series IP address ID 1367 identified as DoS by CESNET experts}
  \label{fig:dos}
\end{figure*}

\paragraph{Anomaly types in dataset} There are many types of anomalies that were described in the literature. All of these anomalies are present in this dataset. Examples of the anomaly types occurring in the dataset are shown in Figure~\ref{fig:anomaly_types}.

The first type of anomaly is \textit{Point anomaly} \cite{chandola2009anomaly}, which is a single data point that significantly deviates from the rest of the datapoints in the time series. There are two types of point anomalies:
\begin{itemize}
    \item Global - A global outlier is a data point that deviates significantly from the overall pattern or distribution of the entire dataset. It is an extreme value when compared to the rest of the data.
    \item Contextual - A data point that is an outlier within a specific context or condition but not necessarily when viewed in a different context.
\end{itemize}

The second type of anomaly is \textit{Collective Anomaly} \cite{chandola2009anomaly}, which is a sequence of datapoints that is anomalous when considered together, even if individual points might not be. There are two types of collective anomalies:
\begin{itemize}
    \item Subsequence - A contiguous subsequence of datapoints that is anomalous compared to the rest of the time series.
    \item Pattern - A sequence of datapoints that together form an unusual pattern, which does not conform to the known patterns of the time series.
\end{itemize}

The third type of anomaly is \textit{Trend Anomaly} \cite{basdekidou2017momentum} which is an unexpected change in the trend of the time series data, such as a sudden shift from a positive to a negative trend. Similar data behaviors in data science are also called data or content drifts \cite{vzliobaite2016overview}, so, sometimes, this anomaly type is called \textit{Drift Anomaly}.

\paragraph{Security incident analysis} In the analysis of the detected anomaly within the dataset, it is crucial to provide a clear explanation of the metrics that characterize the event. Specifically, for the anomaly detected on time series with IP address ID 1367, identified as a Denial of Service (DoS) attack by CESNET experts, the time series metrics were scrutinized in detail, as illustrated in Figure \ref{fig:dos}.

The analysis reveals several key observations. Firstly, both the flow and packet counts increase significantly, yet they rise in parallel, indicating a one-packet flow characteristic. Despite the increase in flows and packets, the number of bytes transmitted does not rise correspondingly, suggesting that the packets involved are small in size. Furthermore, the number of destination ports remains steady, dismissing the possibility of a scan. A significant shift in the TCP/UDP ratio is observed, with almost all the anomalous traffic being UDP, further supporting the DoS identification.

Additional metrics strengthen this conclusion. The traffic direction is near zero, indicating that nearly all the anomalous traffic is directed toward the monitored IP address. The average duration of the flows is also close to zero, which is consistent with one-packet flows typically seen in DoS attacks. The Time To Live (TTL) values remain nearly constant during the anomaly, implying that the traffic likely originates from a single source. Finally, the number of destination IPs shows a slight decrease, reinforcing the idea of a singular sender and confirming the effectiveness of the DoS attack. These combined metrics clearly point to a DoS event targeting the monitored IP address.

\subsection{Validation of usability of dataset}

For the validation of usability, we decided to demonstrate the usage of the dataset's time series for network traffic forecasting. We select an IP address's time series with an ID $103$, the number of IP flows, and a one-hour aggregation interval. To demonstrate time series forecasting applied to the dataset's data, we select the basic SARIMA (Seasonal Autoregressive Integrated Moving Average) as the forecasting model---the order was equal to $(1, 1, 1)$ and seasonal order equal to $(1, 1, 1, 168)$. This means that each of the three components of the model has the same weight in the modeling, and the seasonality was set to one weak (168 hours).

The SARIMA were trained on the monthly data (31 datapoints). We select two different prediction interval to demonstrate the difference. The first prediction window a 7 days long, and the second prediction window is 2 days. Furthermore, the model was retrained with the sliding window equal to the prediction window. Therefore, the model predicted 36 weeks.

\begin{figure*}
  \centering
  \includegraphics[width=\textwidth]{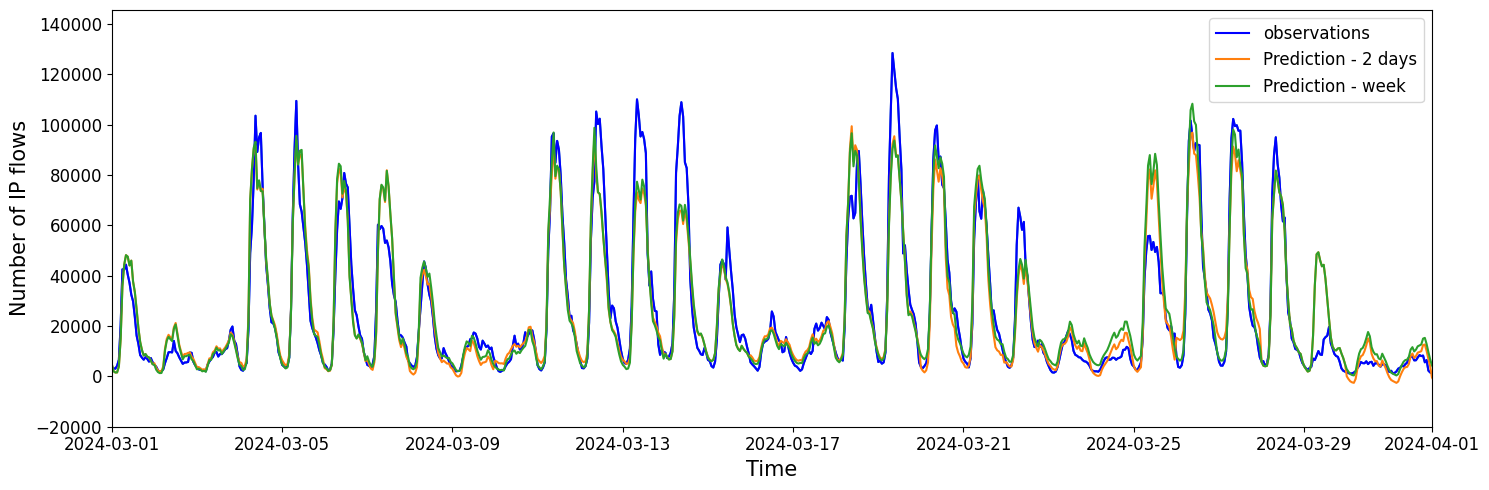}
  \caption{Example of using SARIMA model with order $(1, 1, 1)$ and with seasonal order $(1, 1, 1, 168)$ for forecasting of number of IP flows for time series of IP address's ID 103   }
  \label{fig:sarima_example}
\end{figure*}

The result of this demonstration is shown in Figure \ref{fig:sarima_example}. The Figure contains only one of  The predictions for 2 days, and the model retraining after 2 days resulted in slightly better results, as can be seen. Moreover, we can compare the results by using evaluation metrics like Root Mean Square Error (RMSE), Symmetric Mean Absolute Percentage Error (SMAPE), and $R^2$ Score. In all used metrics, a lower value represents better forecasting performance.  The predicted week data achieves 10951.26 RMSE, 40.66 SMAPE, and 0.77 $R^2$ Score. And the predicted 2 days of data achieve 10293.81 RMSE, 40.86 SMAPE, and 0.79 $R^2$ Score. Therefore, the predicted 2 days achieve better RMSE and $R^2$ Score. However, it achieves a slight decrease in the SMAPE metric.

\section{Usage Notes} 



This section describes recommendations for using the evaluation of network traffic forecasting (and forecasting-based anomaly detection) in detail. We recommend authors follow this evaluation procedure in order to achieve one of the motivations for the creation of this dataset, which is comparability between approaches. We believe that following recommendations will help the community to process our dataset and be able to compare with different approaches that also follow our recommendations. Thus, we present a checklist for addressing all our recommendations which is shown in Table \ref{tab:recommendation-checklist}. Furthermore, the example of using the dataset in the form of a Jupyter Notebook is available on GitHub (\url{https://github.com/koumajos/CESNET-TimeSeries24-Example}). Moreover, the source codes of experiments that provide the evaluation example using our recommendations are also available on this GitHub repository.

\begin{table}[h!]
\centering
\caption{Checklist for Addressing Recommended Evaluation Procedures}
\begin{tabular}{c|p{15cm}}
\textbf{No.} & \textbf{Recommendation} \\ 
\hline
\hypertarget{row1}{(1)} & Specify which dataset(s) are used in the analysis. \\
\rowcolor[HTML]{EFEFEF} \hypertarget{row2}{(2)} & Specify the aggregation interval(s) used. \\ 
\hypertarget{row3}{(3)} & Indicate whether the approach is multivariate or univariate. \\ 
\rowcolor[HTML]{EFEFEF} \hypertarget{row4}{(4)} & Clearly state if not all metrics are used. \\ 
\hypertarget{row5}{(5)} & Document all preprocessing steps, including filtering, normalization, and handling gaps in time series. \\ 
\rowcolor[HTML]{EFEFEF} \hypertarget{row6}{(6)} & Ensure the training phase starts from the beginning of the dataset's time frame (2023-10-09). \\ 
\hypertarget{row7}{(7)} & Specify the duration of the training window. \\ 
\rowcolor[HTML]{EFEFEF} \hypertarget{row8}{(8)} & Define and describe the validation window if employed. \\ 
\hypertarget{row9}{(9)} & Clearly describe the retraining process if the model is retrained during the evaluation phase. \\ 
\rowcolor[HTML]{EFEFEF} \hypertarget{row10}{(10)} & Specify the forecasting horizon (length of time into the future for predictions). \\ 
\rowcolor[HTML]{EFEFEF} \hypertarget{row11}{(11)} & Clearly specify the evaluation metrics used in the article. \\ 
\hypertarget{row12}{(12)} & Provide an overall comparison across each time series using statistical distributions and aggregate statistics. \\ 
\rowcolor[HTML]{EFEFEF} \hypertarget{row13}{(13)} & Assess and document the computational requirements and deployability of the model. \\ 
\hypertarget{row14}{(14)} & Make source codes of your experiments and model publicly available for the community. \\ 
\hline
\end{tabular}
\label{tab:recommendation-checklist}
\end{table}

\paragraph{Dataset Selection}
The dataset utilized in this study is divided into four distinct parts, each of which can be independently used for evaluation: the Full IP address dataset, the Sample IP address dataset, the Institutions dataset, and the Institution 
subnets dataset. Therefore, it is imperative for authors to clearly state which dataset type(s) they are using in their analyses \hyperlink{row1}{(Recommendation 1)}. Moreover, each dataset contains three aggregation intervals; thus, the aggregation interval(s) used must also be specified \hyperlink{row2}{(Recommendation 2)}. In cases where multiple dataset types and/or aggregation levels are employed, results must be reported separately for each dataset and aggregation without combining them. This ensures the clarity and reproducibility of the findings.

Furthermore, the approach must clearly indicate whether it is multivariate (multiple time series metrics are modeled simultaneously) or univariate (each time series metric is modeled individually) \hyperlink{row3}{(Recommendation 3)}. If not all metrics are used, this must be explicitly stated \hyperlink{row4}{(Recommendation 4)}, and comparisons with different approaches should only be made in the following cases. First, for univariate approaches, comparisons should be made individually per each metric. Second, for multivariate approaches, different methods must use the same metrics to ensure accurate comparison.

Moreover, when any preprocessing steps are applied to these dataset parts, it is crucial that these steps are thoroughly described \hyperlink{row5}{(Recommendation 5)}. This includes a detailed description of any filtering, normalization, or transformation processes. Furthermore, this also handles gaps in time series, which must be addressed and described in detail. Particularly, if the preprocessing involves filtering the time series data, this may lead to results that are not directly comparable with studies using unfiltered versions of the same dataset types. Therefore, such preprocessing steps should be clearly justified, and their impact on the analysis should be discussed.

Given that the Full IP address dataset contains more than 275 thousand time series, evaluating methods for all these time series is challenging. Therefore, we encourage authors to create new, smaller datasets featuring interesting time series behaviors from the Full IP address dataset and share them with the community via platforms such as Zenodo. This practice would facilitate further research by providing accessible, focused datasets that highlight specific patterns or anomalies, fostering collaboration and innovation within the research community.

\paragraph{Training Correctness}
To ensure the integrity and validity of the training process, several key guidelines must be followed. The training phase of the time series must always commence from the very beginning of the dataset's time frame, which starts on 2023-10-09 \hyperlink{row6}{(Recommendation 6)}. This causes the model to be trained on the entire range of available data, capturing all relevant trends and patterns, ensuring performance results comparability. The duration of the training window must be explicitly specified in the article \hyperlink{row7}{(Recommendation 7)}. This includes detailing how much historical data is used to train the model before it begins making predictions. 

Moreover, if a validation window is employed during the model development process, the duration and purpose of this validation window must be clearly defined \hyperlink{row8}{(Recommendation 8)}. This helps in assessing the model's performance on unseen data before it is applied to the test set. Furthermore, if the model is retrained during the evaluation phase, this retraining process must be clearly described in the article \hyperlink{row9}{(Recommendation 9)}. Authors should specify the retraining frequency, the data used for retraining, and how the retrained model is validated. 

\paragraph{Forecasting Correctness}
To ensure consistency and transparency in the forecasting methodology, authors must clearly describe the following aspects of their prediction process. Authors must specify the length of time into the future for which predictions are made \hyperlink{row10}{(Recommendation 10)}. This could be a fixed period (e.g., one week ahead) or a rolling window that adjusts over time. Authors should explain whether the window is shifted by a fixed interval or if it adapts based on certain criteria.

\paragraph{Evaluation metrics} The evaluation of the forecasting model should be done by the evaluation metrics. The chosen metrics must be clearly specified in the article \hyperlink{row11}{(Recommendation 11)}. We recommend using the following metrics for evaluation ($n$ is the number of observations, $y_i$ are the actual observed values in the time series, $\hat{y_i} $ are the predicted values):

\begin{itemize}
    \item \textbf{Root Mean Squared Error (RMSE)} calculates the square root of the average squared errors, maintaining the same units as the original data. It is sensitive to large errors, similar to MSE, which helps in detecting significant anomalies. The downside is that this sensitivity might distort the overall model assessment if large errors are not critical. The RMSE can be computed by the equation \ref{rmse}.  \cite{chicco2021coefficient}

    \begin{equation}
    \text{RMSE} = \sqrt{\frac{1}{n} \sum_{i=1}^n (y_i - \hat{y}_i)^2} \label{rmse}
    \end{equation}

    Moreover, for the combination of multiple RMSEs, we recommend using the weighted RMSE, which can be calculated using the equation \ref{rmse-weighted}.

    \begin{equation}
    \text{Weighted RMSE} = \sqrt{\frac{\sum_{i=1}^{n} \sigma_i^2 \cdot \text{RMSE}_i^2}{\sum_{i=1}^{n} \sigma_i^2}}
     \label{rmse-weighted}
    \end{equation}
    
    where $\sigma_i^2$ represents the variance of the true values in the $i^{\text{th}}$ dataset, and $\text{RMSE}_i$ is the corresponding RMSE.

    \item  \textbf{Symmetric Mean Absolute Percentage Error (SMAPE)} addresses some issues of MAPE by symmetrizing the error calculation, providing more stability when actual values are near zero. This symmetry makes it less biased towards overestimations or underestimations. However, SMAPE can still be less intuitive than simpler metrics and may over-penalize certain error types. The SMAPE can be computed by the equation \ref{smape}; the $\epsilon$ in the equation is a small constant that is added to avoid division by zero. \cite{chicco2021coefficient}

    \begin{equation}
    \text{SMAPE} = \frac{100\%}{n} \sum_{i=1}^n \frac{|y_i - \hat{y}_i|}{(|y_i| + |\hat{y}_i| + \epsilon) / 2}  \label{smape}
    \end{equation}

    Moreover, for the combination of multiple SMAPEs, we recommend using the mean and standard deviation of SMAPEs. The weighted SMAPE can also be calculated using the equation \ref{smape-weighted}.

    \begin{equation}
    \text{Weighted SMAPE} = \frac{\sum_{i=1}^{n} \sigma_i^2 \cdot \text{SMAPE}_i}{\sum_{i=1}^{n} \sigma_i^2}
     \label{smape-weighted}
    \end{equation}
        
    where $\sigma_i^2$ represents the variance of the true values in the $i^{\text{th}}$ dataset, and $\text{SMAPE}_i$ is the corresponding SMAPE.

    \item  \textbf{Coefficient of Determination ($R^2$ Score)} measures the proportion of variance explained by the model, providing insight into the model's explanatory power. It is useful for comparing models, but it can be misleading for non-linear models and does not directly measure prediction accuracy. The $R^2$ Score can be computed by the equation \ref{r2score}. \cite{wright1921correlation,chicco2021coefficient}

    \begin{equation}
        R^2 = 1 - \frac{\sum_{i=1}^n (y_i - \hat{y}_i)^2}{\sum_{i=1}^n (y_i - \mu_y)^2}   \label{r2score}
    \end{equation}

    Moreover, for combining multiple $R^2$ scores, we recommend using the weighted $R^2$ score, which can be calculated using the equation \ref{r2score-weighted}.

    \begin{equation}
        \text{Weighted } R^2 = \frac{\sum_{i=1}^{n} \sigma_i^2 \cdot R_i^2}{\sum_{i=1}^{n} \sigma_i^2}
        \label{r2score-weighted}
    \end{equation}
    
    where $\sigma_i^2$ represents the variance of the true values in the $i^{\text{th}}$ dataset, and $R_i^2$ is the corresponding $R^2$ score.

\end{itemize}

\paragraph{Multiple Time Series Evaluation}
All recommended metrics are computed individually for each time series. However, an overall comparison must be made across each time series in the dataset \hyperlink{row12}{(Recommendation  12)}. Therefore, we recommend using statistical distributions per metric to compare overall performance. Initially, aggregate statistics such as the mean and standard deviation should provide a general sense of precision across the dataset, or the weighted variation of the metrics described before can be used. For a more detailed evaluation, we suggest utilizing distribution plots, such as histograms or KDE plots. KDE plots, in particular, are highly effective for detailed comparisons across multiple models, offering a nuanced view of the distribution and performance variations.

\paragraph{Computational Requirements and Deployability of the Model} When evaluating a model's performance, it is crucial not only to consider its precision but also to assess its computational requirements and feasibility of the deployment \hyperlink{row13}{ (Recommendation 13)}. The computational complexity of the model should be analyzed, taking into account factors such as training time, inference speed, and resource consumption (e.g., CPU/GPU usage, memory footprint). Models that require excessive computational resources may be impractical for real-time applications or deployment in environments with limited resources.

\section*{Code Availability}


\begin{table*}
    \centering
    \footnotesize
    \caption{Software used for creating the dataset.}
    \begin{tabular}{l|c|l}
         \textbf{Name} & \textbf{Version} & \textbf{Link}  \\
         \hline
         Ipfixprobe & 4.11.0 & \url{https://github.com/CESNET/ipfixprobe} \\
         \rowcolor[HTML]{EFEFEF} IPFIXcol2 & 2.2.1 & \url{https://github.com/CESNET/ipfixcol2} \\
         NEMEA Framework & 0.14.0 & \url{https://github.com/CESNET/Nemea-Framework} \\
         \rowcolor[HTML]{EFEFEF} NEMEA modules & 2.20.0 & \url{https://github.com/CESNET/Nemea-Modules} \\
         NEMEA Supervisor & 1.8.2 & \url{https://github.com/CESNET/Nemea-Supervisor} \\
         \rowcolor[HTML]{EFEFEF} TimeScaleDB-14 & 2.15.0 & \url{https://www.timescale.com/} \\
         Python  & 3.9.0 & \url{https://www.python.org/downloads/release/python-390/} \\
         \rowcolor[HTML]{EFEFEF} Create Datapoints module & -- & \url{https://github.com/koumajos/CESNET-TimeSeries24-CD} \\
         \hline
    \end{tabular}
    
    \label{tab:softwareversion}
\end{table*}

The dataset has been produced using open-source software. The flow exporter Ipfixprobe, flow collector IPFIXcol2, the NEMEA processing system, and the NEMEA modules are available on GitHub. We use the TimeScaleDB database. Moreover, we provide Create Datapoint module and deployment scripts for the NEMEA Supervisor and for building the database. The versions of used software with links to corresponding repositories are summarized in Table \ref{tab:softwareversion}.

\section*{Acknowledgements} 



This research was funded by the Ministry of Interior of the Czech Republic, grant No. VJ02010024: Flow-Based Encrypted Traffic Analysis and also by the Grant Agency of the CTU in Prague, grant No. SGS23/207/OHK3/3T/18 funded by the MEYS of the Czech Republic. This research was also supported by the Ministry of Education, Youth and Sports of the Czech Republic in the project ``e-Infrastructure CZ'' (LM2023054). Computational resources were provided by the e-INFRA CZ project (ID:90254), supported by the Ministry of Education, Youth and Sports of the Czech Republic.

\section*{Author contributions statement}


J.K. and K.H. propose the time series metrics, J.K. designs and implements the architecture for collecting time series metrics, K.H. and P.Š. design and implement the architecture of capturing network traffic, J.K. and K.H. preprocess data to the final version of the dataset, T.Č. handled funding and supervision, and all authors reviewed the manuscript. 

\section*{Competing interests} 



The authors declare no competing interests.

\bibliographystyle{unsrt}  
\bibliography{references}

\end{document}